%% file: acl_latex.tex
\pdfoutput=1

\documentclass[11pt]{article}

\usepackage[]{acl}

\usepackage{times}
\usepackage{latexsym}
\usepackage{amssymb}

\usepackage[T1]{fontenc}

\usepackage[utf8]{inputenc}

\usepackage{microtype}

\usepackage{inconsolata}
\usepackage{enumitem}
\usepackage{algorithm}
\usepackage{microtype}
\usepackage{boldline}
\usepackage{multirow}
\usepackage{makecell}
\usepackage{hhline}
\usepackage{pgfplots}
\usepackage{tikz}
\usepackage{enumitem}
\usepackage{colortbl}
\usepackage{mathtools}
\usepackage{pifont}
\usepackage{bbm}
\usepackage{amsfonts}
\usepackage{algpseudocode}
\usepackage{booktabs}
\usepackage{xparse}
\usepackage{graphicx}
\usepackage{multicol}
\usetikzlibrary{shapes,arrows}
\usepackage{natbib}
\usepackage{caption}
\usepackage{subcaption}
\usepackage{tikz-dependency}
\usepackage{wrapfig}
\pgfplotsset{compat=1.17} 
\usepackage{subfiles}
\usepackage{readarray}
\usepackage{xcolor}
\usepackage{import}

\input{math_commands.tex}

\usepackage{amsmath}

\title{VP-MEL: Visual Prompts Guided Multimodal Entity Linking}

\author{Hongze Mi$^1$, Jinyuan Li$^2$, Xuying Zhang$^1$, Haoran Cheng$^1$, Jiahao Wang$^1$, Di Sun$^3$, \textbf{Gang Pan$^1$$^{,2,}$\thanks{\hspace{1mm} corresponding author. }}\\
$^1$College of Intelligence and Computing, Tianjin University \\
$^2$School of New Media and Communication, Tianjin University \\
$^3$College of Artificial Intelligence, Tianjin University of Science and Technology\\
  {\tt \{mhzqs, jinyuanli, wjhwtt, pangang\}@tju.edu.cn} \\
}

\begin{document}
\maketitle
\begin{abstract}

Multimodal entity linking (MEL), a task aimed at linking mentions within multimodal contexts to their corresponding entities in a knowledge base (KB), has attracted much attention due to its wide applications. 
However, existing MEL methods primarily rely on mention words as retrieval cues, which limits their ability to effectively utilize both textual and visual information. 
As a result, MEL struggles to retrieve entities accurately, particularly when the focus is on image objects or when mention words are absent from the text.
To solve these issues, we introduce \textbf{V}isual \textbf{P}rompts guided \textbf{M}ultimodal \textbf{E}ntity \textbf{L}inking (VP-MEL). Given a text-image pair, 
VP-MEL links a marked image region (\emph{i.e.}, visual prompt) to its corresponding KB entity.
To support this task, we construct VPWiki, a dataset specifically designed for VP-MEL.
Additionally, we propose the Implicit Information-Enhanced Reasoning (IIER) framework, which enhances visual feature extraction through visual prompts and leverages the pretrained Detective-VLM model to capture latent information. 
Experimental results on VPWiki demonstrate that IIER outperforms baseline methods across multiple benchmarks for VP-MEL.

\end{abstract}

\section{Introduction}
\begin{figure}[h]
  \scriptsize
      \setlength{\belowcaptionskip}{-0.0cm}
    \centering  
    \includegraphics[scale=0.031 ]{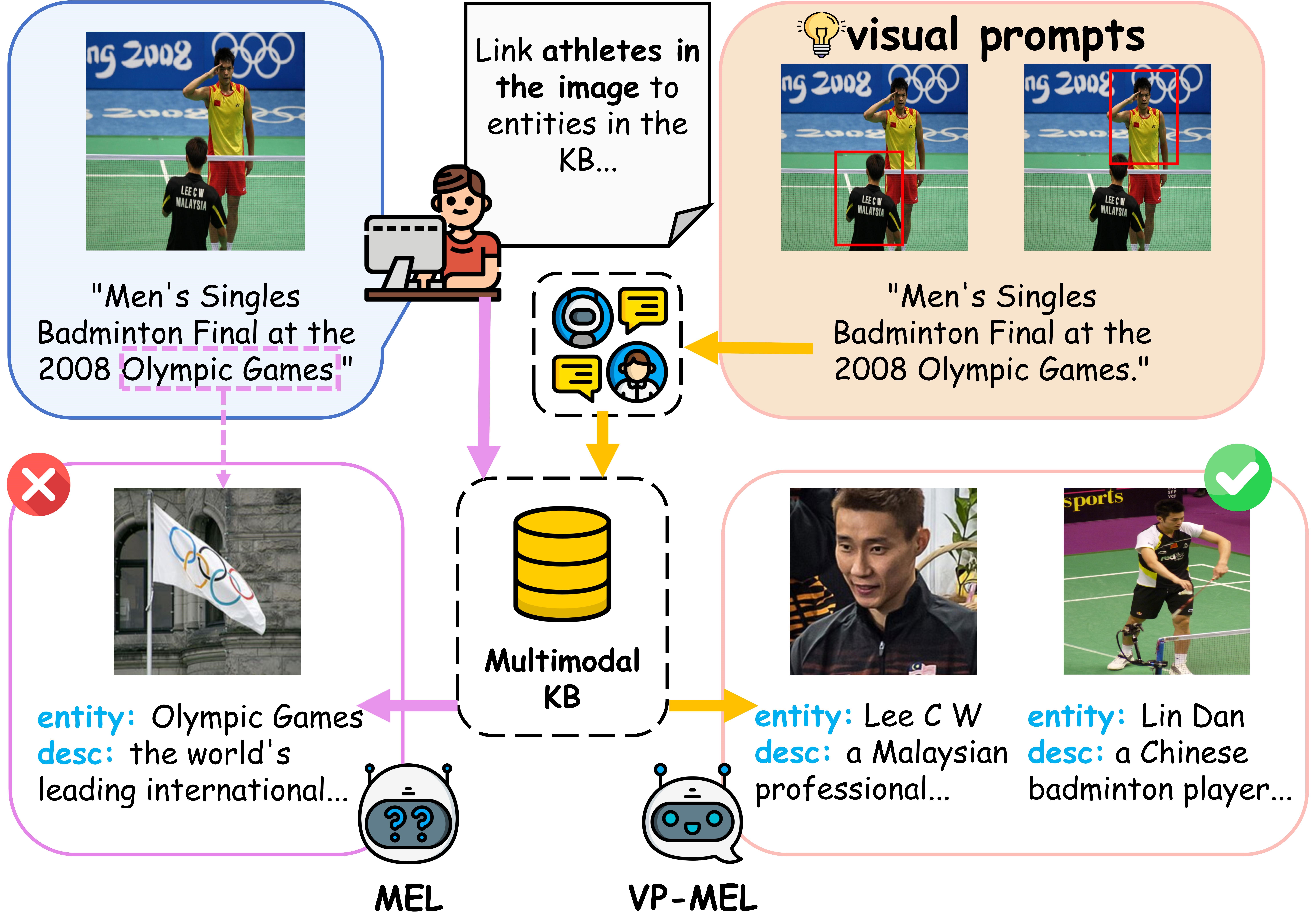}
  \caption{Comparison between MEL and VP-MEL tasks. MEL is typically limited to selecting mentions from text. In contrast, VP-MEL addresses this limitation by using visual prompts to link specific regions in the images to the correct entities in the knowledge base.}
  \label{headphoto}
\end{figure}

Linking ambiguous mentions with multimodal contexts to the referent unambiguous entities in a knowledge base (KB), known as Multimodal Entity Linking (MEL) \citep{moon-etal-2018-multimodal-named}, is an essential task for many multimodal applications. 
Most MEL works \citep{10.1145/3474085.3475400, 10.1145/3477495.3531867, dongjie-huang-2022-multimodal, 10.1145/3580305.3599439, zhang2023multimodal, 10.1145/3581783.3612575, shi-etal-2024-generative} mainly focus on improving the interaction of multimodal information and achieve promising performance. However, existing methods typically represent mentions in the form of mention words and assume that each mention is associated with a high-quality image, which results in two limitations for MEL.\looseness-1

\textbf{Text information dependency:} MEL relies on mention words for entity linking, 
as these words often overlap significantly with entity names in real-world applications. This overlap provides strong cues for identifying entities within the knowledge base (KB). However, MEL performs poorly when mention words are absent or unannotated.
As shown in Figure \ref{headphoto}, without annotated mention words, MEL computes similarity based on the entire text, which can lead to erroneous linking. For instance, it might associate the data with the entity \textit{Olympic Games} due to high textual similarity. Since \textit{Lee C W} and \textit{Lin Dan} are not explicitly mentioned, MEL fails to establish correct links to these entities.
This issue underscores the difficulty of MEL in accurately linking data when mention words related to the entities are missing from the data.\looseness-1

\textbf{Image modality impurity:} Compared to text, images often contain more noise. Misinterpreting or misusing image information can significantly affect the results. Most existing coarse-grained methods \citep{pmlr-v216-yang23d, 10.1145/3580305.3599439, 10.1145/3477495.3531867} directly encode the entire image, making it difficult to eliminate noise interference. 
\citet{song2023dualway} improves MEL performance by extracting fine-grained image information through object detection. However, this approach still relies on sufficient textual information for accurate object localization and is prone to interference from visually similar objects. Therefore, a potentially effective strategy to mitigate image noise is enhancing object localization precision while reducing dependence on textual data.

These limitations hinder the ability of MEL to leverage multimodal data effectively. Images, despite being fundamental to multimodal data, contribute minimally to MEL. Moreover, reliance on textual data constrains MEL performance when text is scarce or incomplete.
So we ask: \emph{Is it possible to link specific objects in multimodal images to the KB even with insufficient textual information? }
Investigating this possibility could unlock the full potential of multimodal data in MEL.

In this paper, we introduce \textbf{V}isual \textbf{P}rompts guided \textbf{M}ultimodal \textbf{E}ntity \textbf{L}inking (VP-MEL), a new task designed for entity linking in image-text pairs, as shown in Figure \ref{headphoto}. 
VP-MEL annotates mentions directly on images using visual prompts,  
eliminating reliance on textual mention words. This approach expands the application scope, enabling effective linking of multimodal data to the KB even when textual information is limited or image-based information is prioritized.
To support this task, we construct the VPWiki dataset based on existing MEL public datasets, where visual prompts are annotated for each mention within the associated images.

To address VP-MEL, we propose the \textbf{I}mplicit \textbf{I}nformation-\textbf{E}nhanced \textbf{R}easoning (IIER) framework.
IIER leverages visual prompts as guiding texture features to focus on specific local image regions. 
To reduce reliance on textual data, it utilizes an external implicit knowledge base to heuristically generate auxiliary information for the reasoning process. 
Specifically, a CLIP visual encoder is employed to extract both global image features and local features guided by visual prompts. AdditionlLy, a Vision-Language Model (VLM) equipped with CLIP visual encoder is pre-trained to generate textual information from visual prompts, supplementing existing text data. IIER provides both supplementary visual and textual information, effectively linking objects in images to the KB.

Main contributions are summarized as follows:
\begin{enumerate}[label=(\roman*)]
    \vspace{-3pt}
    \item
    We introduce VP-MEL, a new entity linking task that replaces traditional mention words with visual prompts, linking specific objects in images to the KB.
    \vspace{-3pt}
    \item
    We construct a high-quality annotated dataset, VPWiki, to enhance the benchmark evaluation of the task. An automated annotation pipeline is proposed to enhance the generation efficiency of the VPWiki dataset.
    \vspace{-3pt}
    \item
    We propose the IIER framework to address VP-MEL task by effectively leveraging multimodal information and reducing reliance on a single modality. Compared to prior methods, IIER achieves a 20\% performance improvement in the VP-MEL task and maintains competitive results in the MEL task.
\end{enumerate}

\section{Related Work}
\subsection{Multimodal Entity Linking}
Given the widespread use of image-text content in social media, the integration of both modalities for entity linking is essential. For example, \citet{moon-etal-2018-multimodal-named} pioneer the use of images to aid entity linking. Building on this, \citet{adjali2020multimodal} and \citet{10.1145/3474085.3475400} construct MEL datasets from Twitter and long movie reviews. Expanding the scope of MEL datasets, \citet{wang-etal-2022-wikidiverse} present a high-quality MEL dataset from Wikinews, featuring diversified contextual topics and entity types. To achieve better performance on these datasets, a multitude of outstanding works in the MEL field \citep{10.1145/3477495.3531867, pmlr-v216-yang23d, 10.1145/3580305.3599439, shi-etal-2024-generative} emerge, focusing on extracting and interacting with multimodal information. \citet{song2023dualway} use object detection to extract visual information from images and better link mention words to correct entities, but still face difficulty in linking images to KBs in the absence of mention words. Although multimodal information can enhance entity linking performance, in these methods, text consistently dominates over images.\looseness-1

\subsection{Vision Prompt}
Region-specific comprehension in complex visual scenes has become a key research topic in the field of Multimodal Computer Vision. Existing methods typically utilize textual coordinate representations \citep{zhu2024minigpt, zhao2023chatspot}, learned positional embeddings \citep{peng2024grounding, zhang2023gpt4roi, zhou2023regionblip}, or Region
of Interest (ROI) features \citep{zhang2023gpt4roi} to anchor language to specific image regions. 
More recently, \citet{10657559} propose a coarse-grained visual prompting solution that directly overlays visual prompts onto the image canvas. In contrast, our VP-MEL provides a fine-grained entity linking method based on visual prompts to reduce reliance on text.

\section{Dataset}
As there is no existing MEL dataset that incorporates visual prompts, annotating a high-quality dataset is crucial to enhance the benchmarking of the VP-MEL task.

\paragraph{Data Collection.}
Our dataset is built on two benchmark MEL datasets, \emph{i.e.}, WikiDiverse \citep{wang-etal-2022-wikidiverse} and WikiMEL \citep{10.1145/3477495.3531867}. 
Appendix \ref{WikiDiverse} provides detailed information. 

\paragraph{Annotation Design.}
Given an image-text pair and the corresponding mention words, annotators need to: 1) annotate relevant visual prompts in the image based on the mention words; 2) remove samples where the image and mention words are unrelated; 3) re-annotate samples with inaccurate pipeline annotations; 4) annotate the entity type for each example (\emph{i.e.}, Person, Organization, Location, Country, Event, Works, Misc).

\paragraph{Annotation Procedure.}
To improve data annotation efficiency, we develop a pipeline that automatically annotates visual prompts in images based on mention words inspired by \citet{li-etal-2024-llms}. In the pipeline, the Visual Entailment Module is employed to evaluate and filter out the highly relevant data. Subsequently, the Visual Grounding Module annotates the visual prompts in the images. The details of the pipeline are provided in Appendix \ref{Pipeline}. The annotation team consists of 10 annotators and 2 experienced experts. All annotators have linguistic knowledge and are instructed with detailed annotation principles. Fleiss Kappa score \citep{fleiss1971measuring} of annotators is 0.83, indicating strong agreement among them. We use the Intersection over Union (IoU) metric to assess annotation quality and discard samples with an IoU score below 0.5.

\begin{figure}[h]
  \scriptsize
      \setlength{\belowcaptionskip}{-0.0cm}
    \centering  
    \includegraphics[scale=0.031]{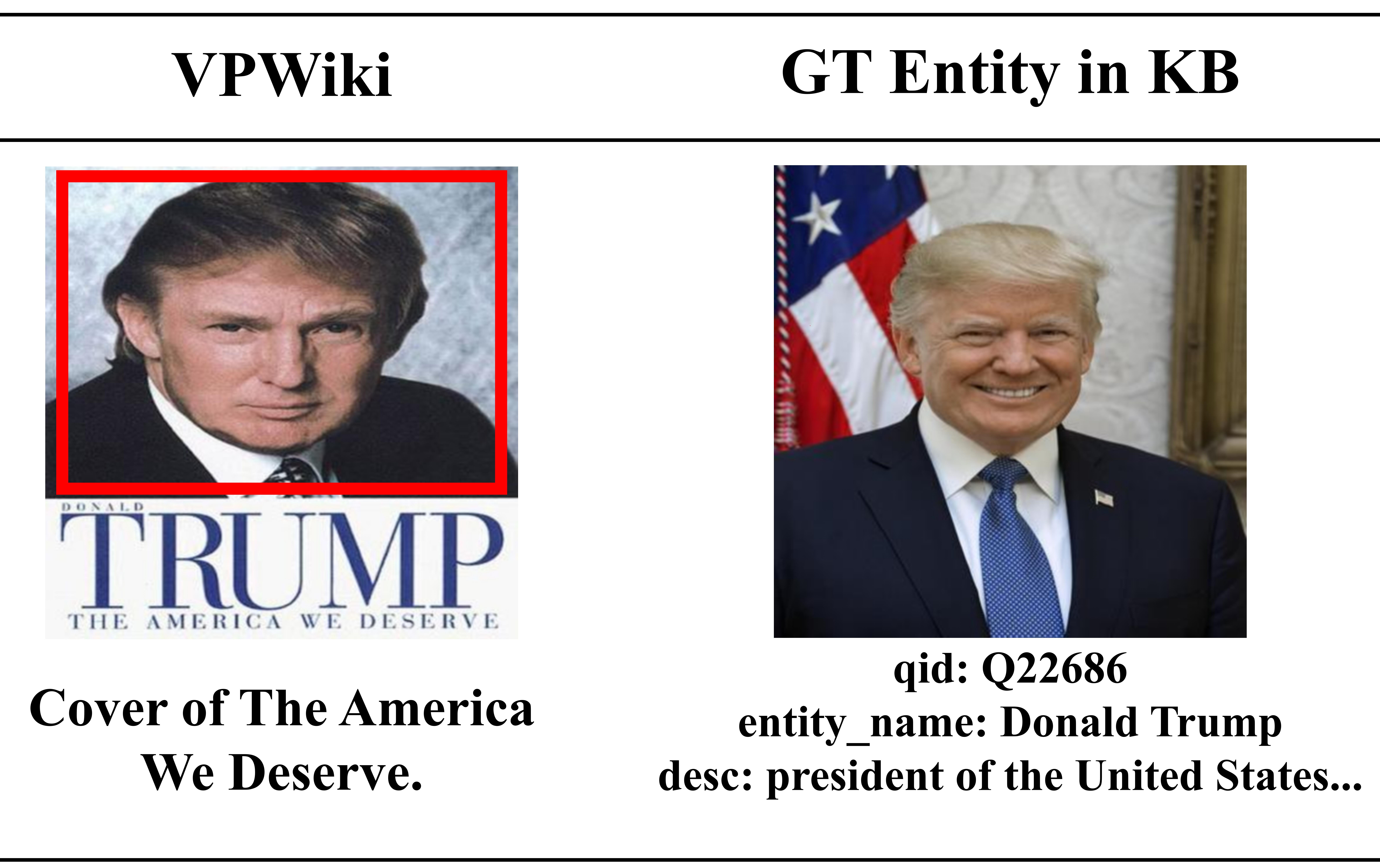}
  \caption{An example from VPWiki. GT denotes the ground truth entity. The red box in the left image represents the visual prompt annotated for the VP-MEL task.}
  \label{VPWiki_example}
\end{figure}

\begin{table}[!]
\small
\setlength\tabcolsep{3pt}
\renewcommand{\arraystretch}{1}
\centering
\begin{tabular}{lcccc}
\toprule

 & \multicolumn{1}{|c}{\textbf{Train}} & \textbf{Dev.} & \textbf{Test}& \textbf{Total}\\
 \midrule 
 pairs & \multicolumn{1}{|c}{8,000} & 1,035& 1,052& 10,087\\
 ment. per pair & \multicolumn{1}{|c}{1.18} &1.16& 1.27&1.19\\ 
 words per pair & \multicolumn{1}{|c}{9.89} &9.80&10.32&9.92\\ 
  \midrule 
\end{tabular}
\caption{Statistics of VPWiki. ment. denotes Mentions.}
\label{tab:Statistics}
\end{table}

\begin{figure}[htbp]
    \begin{minipage}[t]{0.5\linewidth}
        \centering
        \includegraphics[width=\textwidth]{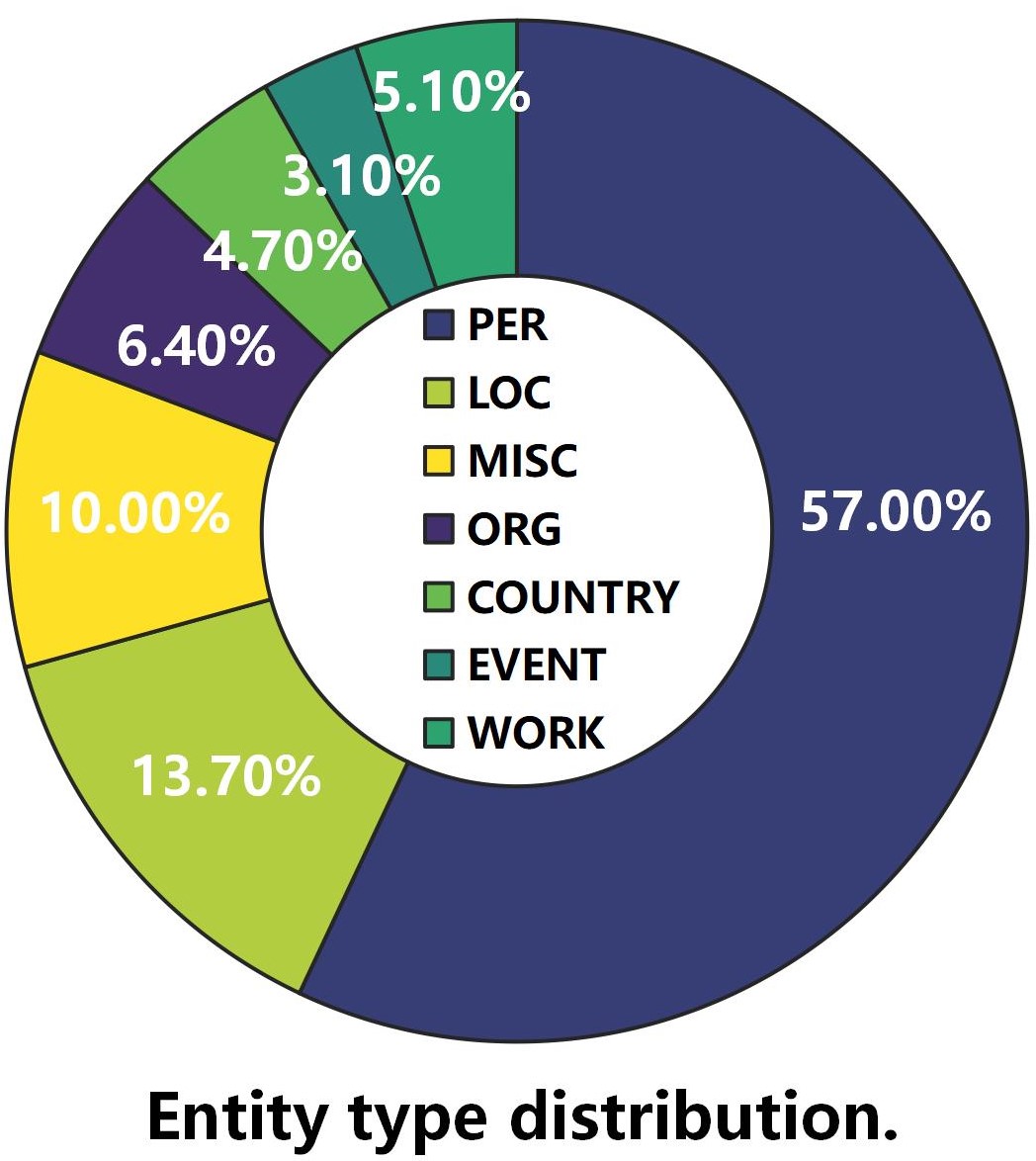}
        \label{More_statistics_l}
        \centerline{(a)}
    \end{minipage}%
    \begin{minipage}[t]{0.5\linewidth}
        \centering
        \includegraphics[width=\textwidth]{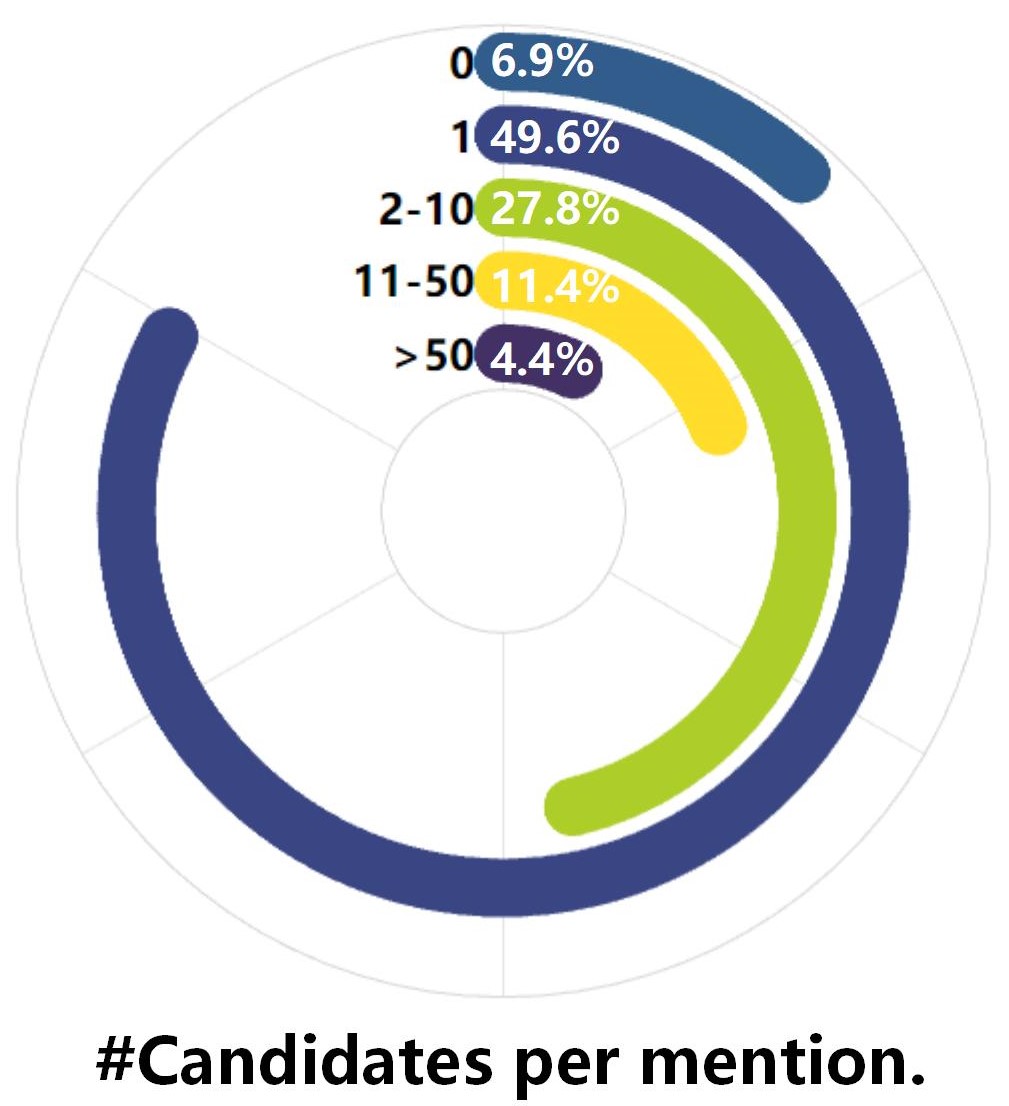}
         \label{More_statistics_r}
        \centerline{(b)}
    \end{minipage}
    \caption{More statistics of VPWiki. (a) Distribution of entity types. (b) Distribution of the number of candidate entities per mention.}
    \label{More_statistics}
\end{figure}

\paragraph{Dataset Analysis.}
Figure \ref{VPWiki_example} illustrates an example from the VPWiki dataset and highlights its differences from traditional MEL data. Additional data samples are provided in \ref{moresamples}. 
The VPWiki dataset comprises a total of 12,720 samples, which are randomly split into training, validation, and test sets with an 8:1:1 ratio.
Detailed statistics of the VPWiki dataset are provided in Table \ref{tab:Statistics}. 
Additionally, Figure \ref{More_statistics} presents the distribution of entity types and the number of candidate entities per mention in the dataset. In Figure \ref{More_statistics}(a), abbreviations are used to represent each entity type. Meanwhile, Figure \ref{More_statistics}(b) shows that as the number of candidate entities per mention increases, the task becomes increasingly challenging.

\section{Task Formulation}
The multimodal knowledge base is constructed by a set of entities $\mathcal{E} = \{E_i\}_{i=1}^N$, where each entity is denoted as $E_i = (e_{n_i}, e_{v_i}, e_{d_i}, e_{a_i})$. Here, 
$e_{n_i}$ denotes the entity name, $e_{v_i}$ represents the entity images, $e_{d_i}$ corresponds to the entity description, and $e_{a_i}$ captures the entity attributes.
A mention is denoted as $M_j = (m_{s_j}, m_{v_j})$, 
where $m_{s_j}$ represents the sentence and $m_{v_j}$ denotes the corresponding image. 
The related entity of mention $M_j$ in the knowledge base is $E_i$.
The objective of the VP-MEL task is to retrieve the ground truth entity $E_i$ from the entity set $\mathcal{E}$ in the knowledge base, based on $M_j$.\looseness-1

\section{Methodology}
\begin{figure*}[t!]
	\centering
	\includegraphics[scale=0.27 ]{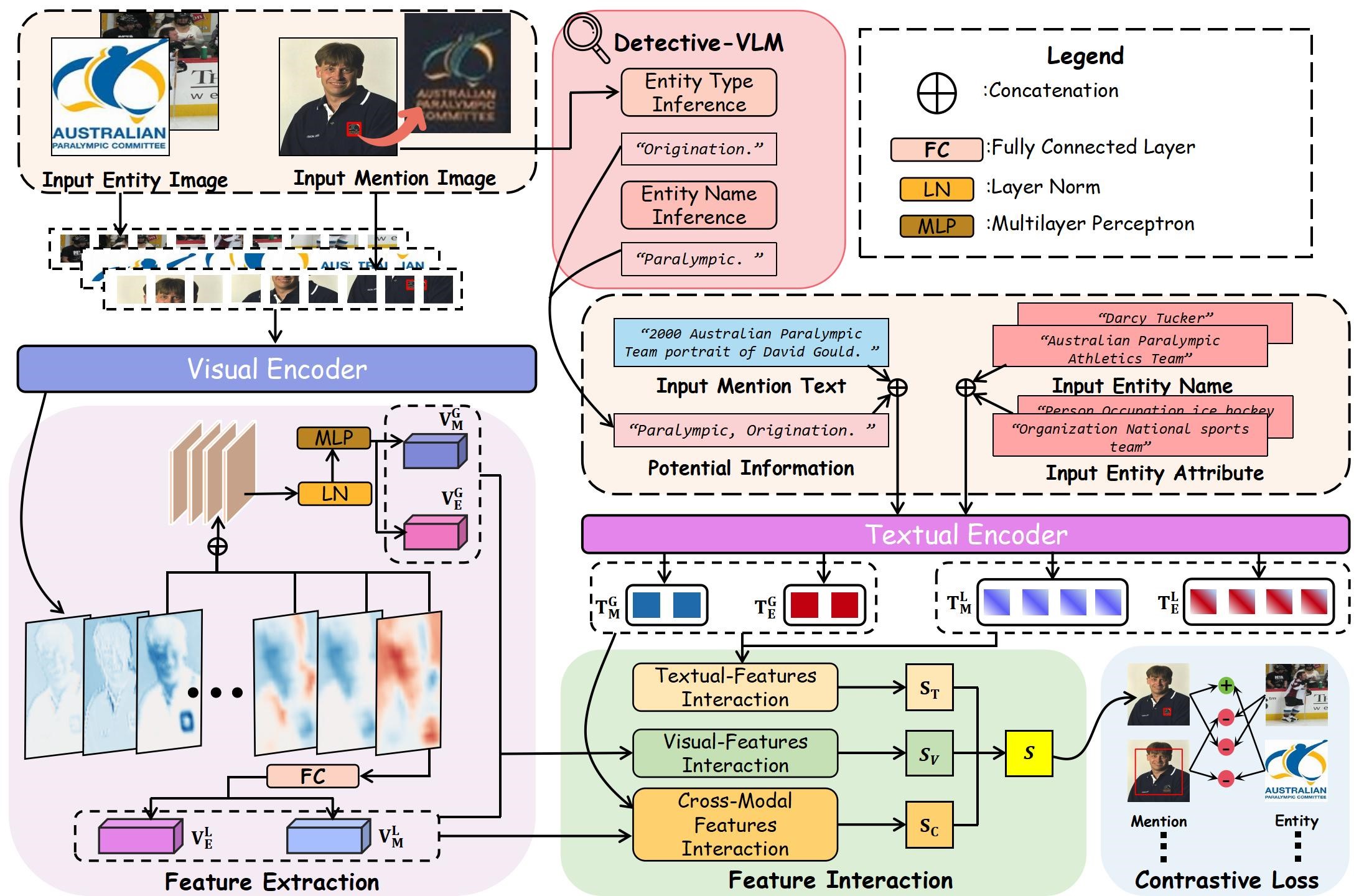}
	\caption{The overall architecture of \textbf{I}mplicit \textbf{I}nformation-\textbf{E}nhanced \textbf{R}easoning (IIER) framework. The image-text pairs of the Mention and Entity are used together as input. Specifically, Mention Text is the sentence corresponding to Mention Image, while Entity Text consists of Entity Name and Entity Attribute corresponding to the Entity Image in the Knowledge Base.}
	\label{fig:architecture}
\end{figure*}

In this section, we describe the proposed IIER framework for the VP-MEL task. 
As illustrated in Figure \ref{fig:architecture}, IIER utilizes visual encoder to extract both deep semantic features and shallow texture features, which are enhanced by visual prompts (§\ref{sec:4.2}). 
To avoid excessive reliance on visual features, the Detective-VLM module is designed to generate supplementary textual information guided by visual prompts(§\ref{sec:4.3}), which is then combined with the original text and processed by the text encoder (§\ref{sec:4.4}). Finally, a similarity score is computed after integrating the visual and textual features(§\ref{sec:4.5}). \looseness-1

\subsection{Visual Encoder}
\label{sec:4.2}
We choose pre-trained CLIP model \citep{dosovitskiy2021an} as our visual encoder. Extensive research \citep{10657559, shtedritski2023does} demonstrates its effectiveness in interpreting visual markers.
The image $m_{v_j}$ of $M_j$ is reshaped into n 2D patches. After this, image patches are processed through visual encoder to extract features. The hidden states extracted from $m_{v_j}$ by the CLIP visual encoder are represented as $V_{M_j}^l = \left[ v_{[CLS]}^0; v_{M_j}^1; v_{M_j}^2;...; v_{M_j}^n \right] \in \mathbb{R}^{(n+1) \times d_c}$, where $d_c$ denotes the dimension of the hidden state and \( l \) denotes the number of layers in the encoder.

Since CLIP focuses on aligning deep features between images and text and may overlook some low-level visual details \citep{zhou2022extract}, we selectively extract features from both the deep and shallow layers of CLIP. Specifically, a shallow feature ($V_{M_j}^3$) is used to represent the textures and geometric shapes in the image, while deep features ($V_{M_j}^{10}$, $V_{M_j}^{11}$, $V_{M_j}^{12}$) are used to represent abstract semantic information. We take the hidden states corresponding to the special [CLS] token ($v_{[CLS]}^0 \in \mathbb{R}^{d_c}$) from these layers as the respective visual features $F^l$. These features are concatenated and normalized using LayerNorm, and then passed through a MLP layer to transform the dimensions to $d_v$, with the output representing the global features of the image $V_{M_j}^G \in \mathbb{R}^{d_v}$.  

\begin{equation}
\mathrm{F^l} = v_{[CLS]}^0 \in V_{M_j}^l, \nonumber
\end{equation}
\begin{equation}
 {V_{M_j}^G}^{'} =
\mathrm{LN}\left(\mathrm{Concat}(F^{3}, F^{10}, F^{11}, F^{12})\right), \nonumber
\end{equation}
\begin{equation}
 V_{M_j}^G =
\mathrm{MLP}\left({V_{M_j}^G}^{'}\right). \nonumber
\end{equation}

Then, hidden states from the output layer of encoder $V_{M_j}^{l}$ are passed through a fully connected layer, which also transforms the dimensions to $d_v$, yielding the local features of the image $V_{M_j}^L \in \mathbb{R}^{(n+1) \times d_v}$:

\vspace{-3pt}
\begin{displaymath}
V_{M_j}^L = \mathrm{FC}\left(V_{M_j}^{l}\right). \nonumber
\end{displaymath}

For the image $e_{v_i}$ of entity $E_i$, the global feature $V_{E_i}^G$ and local feature $V_{E_i}^L$ are obtained using the same method described above.

\subsection{Detective-VLM}
\label{sec:4.3}
Real-world multimodal data often contain challenges such as short texts or image noise. 
In this context, VLMs serve as implicit knowledge bases, can analyze both image and text to infer useful auxiliary information.
Most VLMs \citep{Liu_2024_CVPR, zhu2024minigpt, ye2023mplug, pmlr-v202-li23q} adopt the CLIP visual encoder, enabling them to focus more effectively on markers in images compared to other visual methods \citep{10657559, shtedritski2023does}. Therefore, we instruction fine-tune a VLM to extract effective information from images. The VLM follows template designed below to further mine potential information from the image $m_{v_j}$ and sentence $m_{s_j}$ of mention $M_j$, assisting in subsequent feature extraction:\looseness=-1

\begin{table}[H]
\vspace{-6pt}
    \fontsize{9}{9}\selectfont
    \centerline{
    \begin{tabular}{|l|}
    \hline
    \textbf{Background}: \textcolor{blue}{$\{Image\}$} \\
    \textbf{Text}: \textcolor{blue}{$\{Sentence\}$} \\
    \textbf{Question}: \texttt{Based on the text '\textcolor{blue}{$\{Sentence\}$}', }\\
    \texttt{tell me briefly what is the \textcolor{blue}{$\{Entity\ \allowbreak  Type\}$} and}\\
    \texttt{ \textcolor{blue}{$\{Entity\ \allowbreak  Name\}$} in the red box of the \textcolor{blue}{$\{Image\}$}? }\\
    \textbf{Answer}: \textcolor{blue}{$\{Entity\ \allowbreak Name\}$ $\{Entity\ \allowbreak Type\}$} \\
    \hline
    \end{tabular} {}}
    \vspace{-6pt}
\end{table}
\noindent

We utilize VPWiki dataset to design the fine-tuning dataset, where \textcolor{blue}{$\{Image\}$} and \textcolor{blue}{$\{Sentence\}$} correspond to $m_{v_j}$ and $m_{s_j}$ in $M_j$, respectively. During the inference process, \textcolor{blue}{$\{Entity\ \allowbreak  Name\}$} and \textcolor{blue}{$\{Entity\ \allowbreak  Type\}$} need to be generated by VLM. Details of the dataset and Detective-VLM can be found in Appendix \ref{Detective VLM}.

The objective formula for instruction fine-tuning Detective-VLM is expressed as follows:

\begin{displaymath}
\min_\theta \sum_{i=1}^{N} \mathcal{L}(f_\theta(x_i), y_i),
\end{displaymath}
where \( f \) represents the pre-trained VLM, and \( \theta \) denotes the model parameters. \( N \) represents the number of instruction-output pairs, \( x_i \) is the \( i \)-th instruction, and \( y_i \) is the corresponding desired output. $\mathcal{L}$ is defined as:
\vspace{-3pt}
\begin{displaymath}
\mathcal{L}(f_\theta(x_i), y_i) = -\sum_{t=1}^{T} \log P_\theta(y_i^{(t)} | x_i),
\end{displaymath}
where \( T \) is the length of output sequence, \( y_i(t) \) is the \( t \)-th word of the expected output \( y_i \) at time step \( t \), and \( P \) is conditional probability that the model generates the output \( y_i(t) \) at time step \( t \). 

Detective-VLM aims to ensure that the output is both accurate and relevant, minimizing the likelihood of generating irrelevant information. Notably, we represent the \textbf{Answer} output by VLM as $m_{w_j}$.

\subsection{Textual  Encoder}
\label{sec:4.4}
For the mention $M_j$, after concatenating mention sentence $m_{s_j}$ with $m_{w_j}$, they form the input sequence, with different parts separated by [CLS] and [SEP] tokens:
\vspace{-3pt}
\begin{displaymath}
I_{M_j} = \left[CLS\right]m_{w_j}\left[SEP\right]m_{s_j}\left[SEP\right]. \nonumber
\end{displaymath}

Hidden states of output layer after the input sequence passes through text encoder are represented as $T_{M_j} = \left[ t_{[CLS]}^0; t_{M_j}^1;...; t_{M_j}^{l_t} \right] \in \mathbb{R}^{(l_t+1) \times d_t}$, where $d_t$ represents the dimension of output layer features, and $l_t$ denotes the length of input. We use the hidden state corresponding to [CLS] as the global feature of the text $T_{M_j}^G \in \mathbb{R}^{d_t}$, and the entire hidden states as the local features of the text $T_{M_j}^L \in \mathbb{R}^{(l_t+1) \times d_t}$.

The input sequence for entity $E_i$ consists of the entity name $e_{n_i}$ and entity attributes $e_{a_i}$, which can be represented as:
\vspace{-3pt}
\begin{displaymath}
I_{E_i} = \left[CLS\right]e_{n_i}\left[SEP\right]e_{a_i}\left[SEP\right]. \nonumber
\end{displaymath}

Then, using the above method, we obtain the text features $T_{E_i}^G$ and $T_{E_i}^G$ for the entity.

\subsection{Multimodal Feature Interaction}
\label{sec:4.5}
Inspired by the multi-grained multimodal interaction approach \citep{10.1145/3580305.3599439}, we build the feature interaction part. The multimodal feature interaction section consists of three different units. Notably, this section focuses only on introducing the functions of each unit, detailed mathematical derivations are provided in Appendix \ref{Formula}.

\paragraph{Visual-Features Interaction (VFI).}
Image features of the mention $M_j$ and the entity $E_i$ interact separately. For feature interaction from $M_j$ to $E_i$, after passing through VFI:\looseness=-1
\vspace{-3pt}
\begin{displaymath}
S_{V}^{M2E} = \mathrm{VFI_{M2E}}(V_{M_j}^G, V_{E_i}^G, V_{E_i}^L).
\end{displaymath}

The three input features are sufficiently interacted and integrated, resulting in the similarity matching score $S_{V}^{M2E}$. Similarly, for the feature interaction from $E_i$ to $M_j$, the similarity score $S_{V}^{E2M}$ can be obtained through VFI:
\vspace{-3pt}
\begin{displaymath}
S_{V}^{E2M} = \mathrm{VFI_{E2M}}(V_{E_i}^G, V_{M_j}^G, V_{M_j}^L).
\end{displaymath}

Based on this, the final visual similarity score $S_V$ can be obtained:
\vspace{-3pt}
\begin{displaymath}
S_V = (S_{V}^{M2E} + S_{V}^{E2M}) / 2.
\end{displaymath}

\paragraph{Textual-Features Interaction (TFI).}
TFI computes the dot product of the normalized global features $T_{M_j}^G$ and $T_{E_i}^G$, yielding the text global-to-global similarity score $S_T^{G2G}$:
\vspace{-3pt}
\begin{displaymath}
S_T^{G2G} = T_{M_j}^G \cdot T_{E_i}^G.
\end{displaymath}

To further uncover fine-grained clues within local features, TFI applies attention mechanism to capture context vector from the local features $T_{M_j}^L$ and $T_{E_i}^L$, producing the global-to-local similarity score $S_T^{G2L}$ between the global feature $T_{E_i}^G$ and the context vector:

\begin{displaymath}
S_{T}^{G2L} = \mathrm{TFI_{G2L}}(T_{E_i}^G, T_{M_j}^L, T_{E_i}^L).
\end{displaymath}

Based on this, the final textual similarity score $S_T$ can be obtained:

\begin{displaymath}
S_T = (S_T^{G2G} + S_{T}^{G2L}) / 2.
\end{displaymath}

\paragraph{Cross-Modal Features Interaction (CMFI).}
CMFI performs a fine-grained fusion of features across modalities. It integrates visual and textual features to generate a new context vector, $h_e$:
\vspace{-3pt}
\begin{displaymath}
h_e = \mathrm{CMFI}(T_{E_i}^G, V_{E_i}^L).
\end{displaymath}

The mention is processed similarly to produce the new context vector $h_m$:
\vspace{-3pt}
\begin{displaymath}
h_m = \mathrm{CMFI}(T_{M_j}^G, V_{M_j}^L).
\end{displaymath}

Based on this, the final multimodal similarity score $S_C$ can be obtained:
\vspace{-3pt}
\begin{displaymath}
S_C = h_e \cdot h_m.
\end{displaymath}
 
\subsection{Contrastive Learning}
Based on the three similarity scores $S_V$, $S_T$, and $S_C$, the model is trained using contrastive loss function. For a mention $M$ and entity $E$, the combined similarity score is the average of the similarity scores from the three independent units:
\vspace{-3pt}
\begin{displaymath}
S(M, E) = ( S_V + S_T + S_C ) / 3.
\end{displaymath}

This loss function can be formulated as:
\vspace{-3pt}
\begin{displaymath}
\mathcal{L}_O = -log \frac{\exp(S(M_j, E_i))}{\sum_{i} \exp(S(M_j, E_i^{'}))},
\end{displaymath}
where $E_i$ represents the positive entity corresponding to $M_j$, while $E_i^{'}$ denotes negative entity from the knowledge base $\mathcal{E}$. It is expected to assign higher evaluation to positive mention-entity pairs and lower evaluation to negative ones.

Similarly, the three independent units are trained separately using contrastive loss function:
\vspace{-3pt}
\begin{displaymath}
\mathcal{L}_X = -log \frac{\exp(S_X(M_j, E_i))}{\sum_{i} \exp(S_X(M_j, E_i^{'}))}, X \in \{V, T, C\}.
\end{displaymath}

The final optimization objective function is expressed as:
\vspace{-3pt}
\begin{displaymath}
\mathcal{L} = \mathcal{L}_O + \lambda ( \mathcal{L}_V +\mathcal{L}_T + \mathcal{L}_C),
\end{displaymath}
where $\lambda$ is the hyperparameter to control the loss.

\begin{table*}[ht]
\small
\setlength\tabcolsep{5.4pt}
\renewcommand{\arraystretch}{1.3}
    \centering
    \caption{Performance comparison on the VP-MEL(a) and MEL(b) tasks. Baseline results marked with "$*$" are based on \citet{sui-etal-2024-melov}. Each method is run 5 times with different random seeds, and the mean value of each metric is reported. The best score is highlighted in bold. Detailed evaluation metrics can be found in Appendix \ref{Evaluation Metrics}.}
    \label{tab:combined_tables}
    \begin{subtable}{0.45\textwidth}
        \centering
        \begin{tabular}{lccc}
\toprule
\multirow{2}{*}{\textbf{Methods}}
 & \multicolumn{3}{|c}{\textbf{VP-MEL}}\\
\cline{2-4}
& \multicolumn{1}{|c}{\textbf{H@1}} & \textbf{H@3} & \multicolumn{1}{c}{\textbf{H@5}}\\

  \midrule 

  BLIP-2-xl \citep{pmlr-v202-li23q}  & \multicolumn{1}{|c}{15.86} & 35.41 & \multicolumn{1}{c}{45.32} \\
   BLIP-2-xxl \citep{pmlr-v202-li23q} & \multicolumn{1}{|c}{21.90} & 37.31 & \multicolumn{1}{c}{49.70}\\
  mPLUG-Owl3-7b \citep{ye2023mplug}  & \multicolumn{1}{|c}{ 29.46} & 30.45 & \multicolumn{1}{c}{48.94}\\
  LLaVA-1.5-7b \citep{Liu_2024_CVPR} & \multicolumn{1}{|c}{ 43.20} & 64.35 & \multicolumn{1}{c}{65.71}\\
  LLaVA-1.5-13b \citep{Liu_2024_CVPR} & \multicolumn{1}{|c}{ 32.93} & 65.56 & \multicolumn{1}{c}{66.92}\\
  MiniGPT-4-7b \citep{zhu2024minigpt} & \multicolumn{1}{|c}{ 28.10} & 33.53 & \multicolumn{1}{c}{ 37.31}\\
  MiniGPT-4-13b \citep{zhu2024minigpt} & \multicolumn{1}{|c}{ 37.61} & 37.61 & \multicolumn{1}{c}{40.03}\\

 \midrule 
 
 VELML \citep{zheng2022visual} & \multicolumn{1}{|c}{ 22.51} &  37.61 & \multicolumn{1}{c}{ 43.35}\\
 GHMFC \citep{10.1145/3477495.3531867} & \multicolumn{1}{|c}{ 25.53} & 41.39 & \multicolumn{1}{c}{48.94}\\
 MIMIC \citep{10.1145/3580305.3599439} & \multicolumn{1}{|c}{24.62} & 42.35 & \multicolumn{1}{c}{49.25}\\
 MELOV \citep{song2023dualway} & \multicolumn{1}{|c}{26.44} & 42.75 & \multicolumn{1}{c}{51.51}\\
  
   \midrule 
  \rowcolor[gray]{0.92} IIER(ours) & \multicolumn{1}{|c}{\textbf{48.36}} & \textbf{67.51} & \multicolumn{1}{c}{\textbf{77.50}}\\

\bottomrule
\end{tabular}
\caption{}
\label{BMEL_compare}
    \end{subtable}%
    \hspace{0.08\textwidth} 
    \begin{subtable}{0.45\textwidth}
        \centering
        \begin{tabular}{lccc}
\toprule
\multirow{2}{*}{\textbf{Methods}}
 & \multicolumn{3}{|c}{\textbf{MEL}}\\
\cline{2-4}
& \multicolumn{1}{|c}{\textbf{H@1}} & \textbf{H@3} & \multicolumn{1}{c}{\textbf{H@5}}\\

  \midrule 

 ViLT$^*$ \citep{pmlr-v139-kim21k} & \multicolumn{1}{|c}{34.39} &  51.07 & \multicolumn{1}{c}{ 57.83} \\
  ALBEF$^*$ \citep{li2021align} & \multicolumn{1}{|c}{ 60.59} & 75.59 & \multicolumn{1}{c}{ 83.30} \\
 CLIP$^*$ \citep{radford2021learning} & \multicolumn{1}{|c}{61.21} &  79.63 & \multicolumn{1}{c}{85.18} \\
   METER$^*$ \citep{dou2022empirical} & \multicolumn{1}{|c}{53.14} & 70.93 & \multicolumn{1}{c}{77.59}\\

     \midrule 
  BERT$^*$ \citep{devlin-etal-2019-bert} & \multicolumn{1}{|c}{ 55.77} &  75.73 & \multicolumn{1}{c}{ 83.11}\\
 BLINK$^*$ \citep{wu-etal-2020-scalable} & \multicolumn{1}{|c}{ 57.14} &  78.04 & \multicolumn{1}{c}{ 85.32}\\
  JMEL$^*$ \citep{adjali2020multimodal}  & \multicolumn{1}{|c}{ 37.38} & 54.23 & \multicolumn{1}{c}{ 61.00}\\
  VELML \citep{zheng2022visual} & \multicolumn{1}{|c}{ 55.53} &  78.11 & \multicolumn{1}{c}{ 84.61}\\
  GHMFC \citep{10.1145/3477495.3531867} & \multicolumn{1}{|c}{ 61.17} & 80.53 & \multicolumn{1}{c}{86.21}\\
  MIMIC \citep{10.1145/3580305.3599439} & \multicolumn{1}{|c}{  63.51} &   81.04 & \multicolumn{1}{c}{ 86.43}\\
  MELOV$^*$ \citep{sui-etal-2024-melov} & \multicolumn{1}{|c}{  67.32} &  83.69 & \multicolumn{1}{c}{ 87.54}\\
   \midrule 
  \rowcolor[gray]{0.92} IIER(ours) & \multicolumn{1}{|c}{\textbf{69.47}} & \textbf{84.43} & \multicolumn{1}{c}{\textbf{88.79}}\\

\bottomrule
\end{tabular}
\caption{}
\label{MEL_compare}
    \end{subtable}
\end{table*}

\section{Experiments}
\subsection{Experimental Settings}
All the training and testing are conducted on a device equipped with 4 Intel(R) Xeon(R) Platinum 8380 CPUs and 8 NVIDIA A800-SXM4-80GB GPUs. Detailed experimental settings are provided in Appendix \ref{settings}. To comprehensively evaluate the effectiveness of our approach, we compare IIER with various competitive MEL baselines and VLM baselines. A detailed introduction of these baselines is provided in the Appendix \ref{Baselines}. \looseness-1

For the VP-MEL task experiments, all approaches are evaluated on the VPWiki dataset. And for the MEL task experiments, all approaches are evaluated on the WikiDiverse \citep{wang-etal-2022-wikidiverse} dataset. Additional experiments and detailed explanations are provided in the Appendix \ref{Experiments}.

\subsection{Main Results}

\paragraph{Results on VP-MEL.}

As shown in Table \ref{BMEL_compare}, IIER significantly
outperforms all other methods on VP-MEL task. First, among the VLM methods, LLaVA-1.5 has the smallest performance gap compared to our method, with differences of 5.16\%, 3.16\%, and 11.79\% from IIER across the three metrics, respectively. 
Even so, given the significantly lower training cost compared to LLaVA, IIER offers a clear advantage in efficiency while achieving competitive performance.
Second, there is a notable performance gap between MEL methods and IIER. 
MEL methods struggle with effective entity linking in scenarios where mention words are absent, underscoring their limitations and the robustness of our approach.

\paragraph{Results on MEL.}
Table \ref{MEL_compare} presents the experimental results comparing IIER with other methods on MEL dataset.
During testing, the Detective-VLM analyzes image and text data to generate a concise representation of mention words, which are concatenated with the text and used for entity linking similarity calculation. 
With enhanced visual features and external knowledge, IIER demonstrates excellent performance in the MEL task. Although our work primarily focuses on VP-MEL rather than MEL, IIER still demonstrates strong competitiveness compared to the state-of-the-art MEL method. This highlights the effectiveness of external implicit knowledge in supporting the reasoning process of entity linking.

\begin{table}[!]
\small
\setlength\tabcolsep{4pt}
\renewcommand{\arraystretch}{1.3}
\centering
\begin{tabular}{lccccccc}
\toprule
\multirow{2}{*}{\textbf{Methods}}
 & \multicolumn{3}{|c|}{\textbf{WikiDiverse}}  & \multicolumn{3}{c}{\textbf{WikiDiverse$^*$}}  \\
\cline{2-7}
 & \multicolumn{1}{|c}{\textbf{H@1}} & \textbf{H@3} & \multicolumn{1}{c|}{\textbf{H@5}} & \textbf{H@1} & \textbf{H@3} & \multicolumn{1}{c}{\textbf{H@5}} \\
 \midrule 

 VELML & \multicolumn{1}{|c}{ 55.53} &  78.11 & \multicolumn{1}{c|}{ 84.61} & 15.35 & 26.32 & 31.38\\
 GHMFC & \multicolumn{1}{|c}{61.17} & 80.53 & \multicolumn{1}{c|}{86.21} & 17.37 & 28.97 & 34.36\\
 MIMIC & \multicolumn{1}{|c}{63.51} & 81.04 & \multicolumn{1}{c|}{86.43} & 17.23 & 29.60 & 34.84\\
 MELOV & \multicolumn{1}{|c}{67.32} & 83.69 & \multicolumn{1}{c|}{87.54} & 17.66 & 30.03 & 36.43 \\
  \midrule 
    \rowcolor[gray]{0.92} IIER & \multicolumn{1}{|c}{\textbf{69.47}} & \textbf{84.43}& \multicolumn{1}{c|}{\textbf{88.79}} & \textbf{23.87} &\textbf{38.37} &\multicolumn{1}{c}{\textbf{45.14}}\\
    \bottomrule
\end{tabular}
\caption{Performance comparison in the absence of mention words on the WikiDiverse dataset. The symbol "$*$" represents the dataset without annotated mention words.\looseness-1}
\label{tab:Mention words result}
\end{table}

\subsection{Detailed Analysis}
\paragraph{Influence of Mention Words on MEL Methods.}
As shown in Table \ref{tab:Mention words result}, 
the performance of MEL methods drop significantly across all three metrics in the absence of mention words. The average performance decline is 72.65\%, 64.48\%, and 60.28\%, respectively.
This indicates that MEL methods fail to extract meaningful information from visual and textual data, making them unsuitable for tasks without mention words.
In contrast, even without Detective-VLM, visual prompts, or mention words, IIER can still achieve  the best metrics. This demonstrates that IIER in the VP-MEL task possesses a stronger capability to leverage both image and text  information effectively.

\begin{table}[!]
\small
\setlength\tabcolsep{2.5pt}
\renewcommand{\arraystretch}{1.3}
\centering
\begin{tabular}{lcccccc}
\toprule
\multirow{2}{*}{\textbf{Methods}}
 & \multicolumn{5}{|c}{VP-MEL}\\
\cline{2-6}
 & \multicolumn{1}{|c}{\textbf{H@1}} & \textbf{H@3}& \textbf{H@5}& \textbf{H@10} & \multicolumn{1}{c}{\textbf{H@20}}\\
 \midrule 
 MiniGPT-4-7b & \multicolumn{1}{|c}{28.55} & 43.66 & 52.27& 62.99 &70.70\\
 MiniGPT-4-13b & \multicolumn{1}{|c}{27.04} & 43.96 & 53.02 & 63.44 &70.72\\
 BLIP-2-xl  & \multicolumn{1}{|c}{37.16} & 54.38 & 59.52& 66.62 &72.81\\
 BLIP-2-xxl  & \multicolumn{1}{|c}{40.63} & 54.53 & 61.78& 68.73 &74.62\\
 LLaVA-1.5-7b  & \multicolumn{1}{|c}{42.45} & 63.14 & 69.03& 76.74 &82.33\\
 LLaVA-1.5-13b  & \multicolumn{1}{|c}{41.54} & 59.37 & 66.92& 73.11 &77.80\\
  \midrule 
  \rowcolor[gray]{0.92} Detective-VLM(ours) & \multicolumn{1}{|c}{\textbf{48.36}} & \textbf{67.51} & \textbf{77.50}& \textbf{82.59} &\textbf{87.90}\\
  \bottomrule
\end{tabular}
\caption{Performance comparison in different VLMs.}
\label{tab:VLM result}
\end{table}

\paragraph{Effect Analysis of Detective-VLM.}
As shown in Table \ref{tab:VLM result}, we evaluate the effectiveness of Detective-VLM by replacing it with various VLMs and analyzing the results. Our method achieves the best performance across all metrics.  In particular, Detective-VLM shows an absolute improvement of 5.91\% in Hit@1 compared to the second-best approach.
In contrast, non-fine-tuned VLMs often produce a large amount of irrelevant information, which hampers subsequent processing.

\begin{table}[!]
\small
\setlength\tabcolsep{2pt}
\renewcommand{\arraystretch}{1.3}
\centering
\begin{tabular}{lcccccc}
\toprule
\multirow{2}{*}{\textbf{$V^G$-Layer}}
 & \multicolumn{5}{|c}{VP-MEL}\\
\cline{2-6}
 & \multicolumn{1}{|c}{\textbf{H@1}} & \textbf{H@3}& \textbf{H@5}& \textbf{H@10} & \multicolumn{1}{c}{\textbf{H@20}}\\
 \midrule 
 Single Shallow Layer & \multicolumn{1}{|c}{39.88} & 60.88 & 71.00 & 79.31 &86.56\\
 Single Deep Layer  & \multicolumn{1}{|c}{39.73} & 58.91 & 69.94& 80.82 &\textbf{88.07}\\
  \midrule 
 (Shallow+Deep) Layers  & \multicolumn{1}{|c}{43.66} & 60.88 & 68.58& 78.70 &84.29\\
 (3 Shallow+Deep) Layers  & \multicolumn{1}{|c}{40.33} & 59.22 & 67.37& 75.38 &83.23\\ 

  \midrule 
  \rowcolor[gray]{0.92} IIER & \multicolumn{1}{|c}{\textbf{48.36}} & \textbf{67.51} & \textbf{77.50}& \textbf{82.59} &87.90\\
  \bottomrule
\end{tabular}
\caption{Performance comparison across different feature layers in $V^G$.}
\label{tab:Visual Feature}
\end{table}

\paragraph{Contributions of Visual Features from Different Layers.}
As shown in Table \ref{tab:Visual Feature}, we combine visual features from different layers during the extraction of $V^G$ to compare the effects of various combinations. In the deeper layers of CLIP visual encoder, the model tends to focus more on abstract, high-level concepts. VP-MEL focuses on aligning high-level concepts between images and text, facilitating the capture of their semantic correspondence. This explains why using a single deep layer feature achieves the highest H@20 score of 88.07\%. However, in the VP-MEL task, low-level texture details are equally important. Shallow texture features need to be extracted to help the model focus on the presence of visual prompts. Based on this, we choose to concatenate the deep features with the shallow features. Experimental results show that the best performance is achieved when the proportion of deep features is larger.

\begin{table}[!]
\small
\setlength\tabcolsep{6.5pt}
\renewcommand{\arraystretch}{1.5}
\centering
\begin{tabular}{lcccccc}
\toprule
\multirow{2}{*}{\textbf{Methods}}
 & \multicolumn{5}{|c}{VP-MEL}\\
\cline{2-6}
 & \multicolumn{1}{|c}{\textbf{H@1}} & \textbf{H@3}& \textbf{H@5}& \textbf{H@10} & \multicolumn{1}{c}{\textbf{H@20}}\\
 \midrule 
 IIER & \multicolumn{1}{|c}{\textbf{48.36}} & \textbf{67.51} & \textbf{77.50}& \textbf{82.59} &\textbf{87.90}\\
  \midrule 
 IIER$^\dag$  & \multicolumn{1}{|c}{35.65} & 53.93 & 65.26& 73.57 &80.51\\
 IIER$^*$  & \multicolumn{1}{|c}{35.03} & 53.80 & 65.01& 73.26 &80.39\\ 

  \midrule 
\end{tabular}
\caption{The model marked "$\dag$" without VLM. The model marked "$*$" without VLM and Visual Prompts.}
\label{tab:Ablation}
\end{table}

\paragraph{Ablation Study.}
In Table \ref{tab:Ablation}, we conduct ablation
study on the IIER framework. 
First, we remove the Detective-VLM module from IIER, which results in a decline across all metrics. Notably, even without Detective-VLM, IIER shows robust entity linking performance, outperforming MEL methods as shown in Table \ref{BMEL_compare}. 
This highlights the ability of IIER to efficiently leverage multimodal information from both images and text.
Subsequently, removing the visual prompts from the images results in a decline across all metrics, emphasizing the crucial role of visual prompts in guiding the model to focus on relevant regions within the images. 
Note that the slight decrease in metrics does not suggest a diminished significance of visual prompts, as they are integral to the functioning of the VLM.

 \section{Conclusion}
In this paper, we introduce a new task, VP-MEL, aiming to link visual regions in image-text pairs to the corresponding entities in knowledge base, guided by visual prompts. To support this task, we construct a high-quality dataset, VPWiki, leveraging an automated annotation pipeline to enhance the efficiency of data annotation. To address the VP-MEL task, we propose IIER, 
a framework that can effectively utilize visual prompts to extract enhanced local visual features and generate supplementary textual information.
IIER maintains a balance between visual and textual features, preventing excessive reliance on a single modality. Extensive experimental results demonstrate that IIER surpasses state-of-the-art methods. Moreover, VP-MEL significantly alleviates the constraints of mention words and expands the applicability of MEL to real-world scenarios.

\section*{Limitations}
VP-MEL expands the application scenarios of MEL, allowing users to directly annotate areas of interest within images. However, this requires the image and text to be correlated to some extent. When the image and text are not correlated, the performance of VP-MEL can be limited. We choose to use grounding box as the format for visual prompts. In real-world application scenarios, users may use any irregular shape to make markings. In future work, the design of visual prompts will be enhanced. We hope this work can encourage more research to apply the recent advanced techniques from both natural language processing and computer vision fields to improve its performance.

\section*{Ethics Statement}
The datasets employed in this paper, WikiDiverse, WikiMEL, and RichpediaMEL, are all publicly accessible. As such, the images, texts, and knowledge bases referenced in this study do not infringe upon the privacy rights of any individual.


\bibliography{cite}

\appendix
\section{Appendix}

\subsection{Experimental Settings}
\label{settings}
For our proposed model framework, we use pre-trained ViT-B/32 \citep{dosovitskiy2021an} as the visual encoder, initialized with weights from CLIP-ViT-Base-Patch32\footnote{\url{https://huggingface.co/openai/clip-vit-base-patch32}}, with \( d_v \) and \( d_c \) set to 96. The number of epochs is set to 20, and the learning rate is tuned to $1 \times 10^{-5}$. The batch size is set to 128. In the loss function, $\lambda$ is set to 1. For the text encoder, we select pre-trained BERT model \citep{devlin-etal-2019-bert}, setting the maximum input length for text to 40 and the output feature dimension \( d_t \) to 512. Without including the VLM, the size of the trainable parameters is 153 M, and the total estimated model parameters size is 613 M. We train and test on a device equipped with 4 Intel(R) Xeon(R) Platinum 8380 CPUs and 8 NVIDIA A800-SXM4-80GB GPUs.

\subsection{Descriptions of Baselines}
\label{Baselines}
To thoroughly evaluate the performance of our method, we compare it against strong MEL baselines, including \textbf{BERT} \citep{devlin-etal-2019-bert}, \textbf{BLINK} \citep{wu-etal-2020-scalable}, \textbf{JMEL} \citep{adjali2020multimodal}, \textbf{VELML} \citep{zheng2022visual}, \textbf{GHMFC} \citep{10.1145/3477495.3531867}, \textbf{MIMIC} \citep{10.1145/3580305.3599439} and \textbf{MELOV} \citep{sui-etal-2024-melov}.

Additionally, we select robust VLMs for comparison, including \textbf{BLIP-2-xl} \footnote{\url{https://huggingface.co/Salesforce/blip2-flan-t5-xl-coco}}, \textbf{BLIP-2-xxl} \footnote{\url{https://huggingface.co/Salesforce/blip2-flan-t5-xxl}} \citep{pmlr-v202-li23q}, \textbf{mPLUG-Owl3-7b} \footnote{\url{https://huggingface.co/mPLUG/mPLUG-Owl3-7B-240728}} \citep{ye2023mplug}, \textbf{LLaVA-1.5-7b} \footnote{\url{https://huggingface.co/liuhaotian/llava-v1.5-7b}}, \textbf{LLaVA-1.5-13b} \footnote{\url{https://huggingface.co/liuhaotian/llava-v1.5-13b}} \citep{Liu_2024_CVPR}, \textbf{MiniGPT-4-7b} \footnote{\url{https://drive.google.com/file/d/1RY9jV0dyqLX-o38LrumkKRh6Jtaop58R/view?usp=sharing}}, \textbf{MiniGPT-4-13b} \footnote{\url{https://drive.google.com/file/d/1a4zLvaiDBr-36pasffmgpvH5P7CKmpze/view?usp=share_link}} \citep{zhu2024minigpt}, \textbf{ViLT} \citep{pmlr-v139-kim21k}, \textbf{ALBEF} \citep{li2021align}, \textbf{CLIP} \citep{radford2021learning}, and \textbf{METER} \citep{dou2022empirical}. We reimplemented JMEL, VELML and MELOV according to the original literature due to they did not release the code.  We ran the official implementations of the other baselines with their default settings.

\noindent\textbullet \textbf{BERT} \citep{devlin-etal-2019-bert} is a pre-trained language model based on the Transformer architecture, designed to deeply model contextual information from both directions of a text, generating general-purpose word representations.

\noindent\textbullet \textbf{BLINK} \citep{wu-etal-2020-scalable} present a two-stage zero-shot linking algorithm, where each entity is defined only by a short textual description.

\noindent\textbullet \textbf{JMEL} \citep{adjali2020multimodal} extracts both unigram and bigram embeddings as textual features. Different features are fused by concatenation and
a fully connected layer. 

\noindent\textbullet \textbf{VELML} \citep{zheng2022visual} utilizes VGG-16 network to obtain object-level visual features. The two modalities are fused with additional attention mechanism.

\noindent\textbullet \textbf{GHMFC} \citep{10.1145/3477495.3531867} extracts hierarchical features of text and visual co-attention through the multi-modal co-attention mechanism.

\noindent\textbullet \textbf{MIMIC} \citep{10.1145/3580305.3599439} devise three interaction units to sufficiently explore and extract diverse multimodal interactions and patterns for entity
linking.

\noindent\textbullet \textbf{MELOV} \citep{sui-etal-2024-melov} incorporates inter-modality generation
and intra-modality aggregation.

\noindent\textbullet \textbf{BLIP-2} \citep{pmlr-v202-li23q}  effectively utilizes the noisy web data by bootstrapping the captions, where a captioner generates synthetic captions and a filter removes the noisy ones.

\noindent\textbullet \textbf{mPLUG-Owl3} \citep{ye2023mplug} propose novel hyper attention blocks to efficiently integrate vision and language into a common language-guided semantic space, thereby facilitating the processing of extended multi-image scenarios. 

\noindent\textbullet \textbf{LLaVA-1.5} \citep{Liu_2024_CVPR} is an end-to-end trained
 large multimodal model that connects a vision encoder and an LLM for general purpose visual and language understanding.

\noindent\textbullet \textbf{MiniGPT-4} \citep{zhu2024minigpt} aligns a frozen visual encoder with a frozen LLM, Vicuna, using just one projection layer.

\noindent\textbullet \textbf{ViLT} \citep{pmlr-v139-kim21k} commissions the transformer module to extract and process visual features in place of a separate deep visual embedder.

\noindent\textbullet \textbf{ALBEF} \citep{li2021align} introduce a contrastive loss to align the image and text representations before fusing them through cross-modal attention, which enables more grounded vision and language representation learning.

\noindent\textbullet \textbf{CLIP} \citep{radford2021learning} is a neural network trained on a variety of (image, text) pairs. It can be instructed in natural language to predict the most relevant text snippet, given an image.

\noindent\textbullet \textbf{METER} \citep{dou2022empirical} systematically investigate how to train a fully-transformer VLP model in an end-to-end manner.

\subsection{Evaluation Metrics}
\label{Evaluation Metrics}
For evaluation, we utilize Top-k accuracy as the metric that can be calculated by the following formula:\looseness-1
\begin{displaymath}
\mathrm{Accuracy}_{top-k} = \frac{1}{N} {\sum_{i}^N I(t_i \in y_i^k)},
\end{displaymath}
where \( N \) represents the total number of samples, and \( I \) is the indicator function. When the receiving condition is satisfied, \( I \) is set to
1, and 0 otherwise.

\subsection{Detective-VLM}
\label{Detective VLM}
 Detective-VLM is based on the mplug-owl2 framework \citep{Ye_2024_CVPR}, with instruction fine-tuning carried out using the mplug-owl2-llama2-7b model \footnote{\url{https://huggingface.co/MAGAer13/mplug-owl2-llama2-7b}}. \looseness-1
 
 We utilize VPWiki dataset to design the fine-tuning dataset, where \textcolor{blue}{$\{Image\}$} and \textcolor{blue}{$\{Sentence\}$} correspond to $m_{v_j}$ and $m_{s_j}$ in $M_j$, respectively. In the fine-tuning dataset, the \textcolor{blue}{$\{Entity\ \allowbreak  Name\}$} corresponds to the mention words in $M_j$ that are associated with the Visual prompt, the \textcolor{blue}{$\{Entity\ \allowbreak  Type\}$} is one of [$Person$, $Organization$, $Location$, $Country$, $Event$, $Works$, $Misc$].

\subsection{Feature Interaction Formula}
\label{Formula}
\paragraph{Visual-Features Interaction (VFI).}
The two similarity scores $S_{V}^{M2E}$ and $S_{V}^{E2M}$ in visual feature interaction are calculated using the same method. Here, we take $S_{V}^{M2E}$ as an example.

\begin{displaymath}
\overline{h}_p = \mathrm{MeanPooling} (V_{E_i}^L),
\end{displaymath}
\begin{displaymath}
h_{vc} = \mathrm{FC} (\mathrm{LayerNorm}(\overline{h}_p + V_{M_j}^G)),
\end{displaymath}
\begin{displaymath}
h_{vg} = \mathrm{Tanh} (\mathrm{FC}(h_{vc})),
\end{displaymath}
\begin{displaymath}
h_{v} = \mathrm{LayerNorm} (h_{vg} * h_{vc} + V_{E_i}^G),
\end{displaymath}
\begin{displaymath}
S_{V}^{M2E} = h_{v} \cdot V_{M_j}^G.
\end{displaymath}

\paragraph{Textual-Features Interaction (TFI).}
The calculation of the global-to-local similarity score $S_{T}^{G2L}$ incorporates an attention mechanism as follows:
\begin{displaymath}
Q,K,V = T_{E_i}^LW_{tq}, T_{M_j}^LW_{tk}, T_{M_j}^LW_{tv},
\end{displaymath}
\begin{displaymath}
H_{t} = \mathrm{softmax}(\frac{QK^T} {\sqrt{d_T}})V,
\end{displaymath}
where $T_{E_i}^LW_{tq}$, $T_{M_j}^LW_{tk}$, $T_{M_j}^LW_{tv}$ are learnable matrices.
\begin{displaymath}
h_{t} = \mathrm{LayerNorm} (\mathrm{MeanPooling}(H_t)),
\end{displaymath}
\begin{displaymath}
S_{T}^{G2L} = \mathrm{FC} (T_{E_i}^G) \cdot h_t.
\end{displaymath}

\paragraph{Cross-Modal Features Interaction (CMFI).}
CMFI performs alignment and fusion of features from different modalities. 
\begin{displaymath}
h_{et}, h_{mt} = \mathrm{FC}_{c1} (T_{E_i}^G), \mathrm{FC}_{c1} (T_{M_j}^G),
\end{displaymath}
\begin{displaymath}
H_{ev}, H_{mv} = \mathrm{FC}_{c2} (V_{E_i}^L), \mathrm{FC}_{c2} (V_{M_j}^L),
\end{displaymath}
in which ${FC}_{c1}$ is defined by $W_{c1} \in \mathbb{R}^{d_t \times d_c}$ and $b_{c1} \in \mathbb{R}^{d_c}$,  ${FC}_{c2}$ is defined by $W_{c2} \in \mathbb{R}^{d_v \times d_c}$ and $b_{c2} \in \mathbb{R}^{d_c}$.
\begin{displaymath}
\alpha_i =  \frac{exp(h_{et} \cdot H_{ev}^i)}{\sum_{1}^{n+1} exp(h_{et} \cdot H_{ev}^i)},
\end{displaymath}
\begin{displaymath}
h_{ec} = \sum\limits_{i}^{n} \alpha_i * H_{ev}^i, i \in [1,2,...,(n+1)],
\end{displaymath}
\begin{displaymath}
h_{eg}= \mathrm{Tanh}(\mathrm{FC}_{c3} (h_{et}),
\end{displaymath}
in which ${FC}_{c3}$ is defined by $W_{c3} \in \mathbb{R}^{d_c \times d_c}$ and $b_{c3} \in \mathbb{R}^{d_c}$.
\begin{displaymath}
h_{e}= \mathrm{ LayerNorm}(h_{eg}*h_{et} + h_{ec}).
\end{displaymath}

By replacing inputs \( h_{et} \) and \( H_{ev} \) with \( h_{mt} \) and \( H_{mv} \), \( h_m \) can be obtained using the aforementioned formula.

\subsection{Data Annotation Pipeline}
\label{Pipeline}
Please note that the pipeline serves as a preprocessing stage for data annotation. We use the Visual Entailment Module and the Visual Grounding Module to automatically annotate visual prompts in the images. While the accuracy of the pipeline is limited—such as its difficulty in distinguishing between specific individuals when multiple people are present—it still plays a crucial role in improving annotation efficiency. Due to these limitations, manual verification and re-annotation are necessary after pipeline processing. However, for annotators, making a simple "yes or no" judgment is much easier than selecting a specific individual. As a result, even with limited accuracy, the pipeline significantly boosts the overall efficiency of the annotation process.

For the Visual Entailment Module and Visual Grounding Module, we choose $\mathrm{OFA}_{large(VE)}$ and $\mathrm{OFA}_{large(VG)}$ \citep{wang2022ofa}, respectively. 

\begin{table}[!]
\small
\setlength\tabcolsep{4pt}
\renewcommand{\arraystretch}{1.3}
\centering
\begin{tabular}{lcccc}
\toprule

 & \multicolumn{1}{|c|}{\textbf{WIKIDiverse}} & \textbf{WikiMEL}& \multicolumn{1}{|c}{\textbf{RichpediaMEL}}\\
 \midrule 
 Sentences & \multicolumn{1}{|c|}{7,405} & 22,070& \multicolumn{1}{|c}{17,724}\\
 M. in train & \multicolumn{1}{|c|}{11,351} & 18,092 & \multicolumn{1}{|c}{12,463}\\
 M. in valid & \multicolumn{1}{|c|}{1,664} &2,585 &\multicolumn{1}{|c}{1,780}\\ 
 M. in test & \multicolumn{1}{|c|}{2,078} &5,169 &\multicolumn{1}{|c}{3,562}\\ 
 Entities & \multicolumn{1}{|c|}{132,460} &109,976 &\multicolumn{1}{|c}{160,935}\\ 

  \midrule 
\end{tabular}
\caption{Statistics of WIKIDiverse, WikiMEL, and RichpediaMEL. M. denotes Mentions.}
\label{tab:WIKIDiverse}
\end{table}

\begin{figure}[h]
  \scriptsize
      \setlength{\belowcaptionskip}{-0.0cm}
    \centering  
    \includegraphics[scale=0.6 ,trim={2cm 2cm 2cm 2cm}, clip]{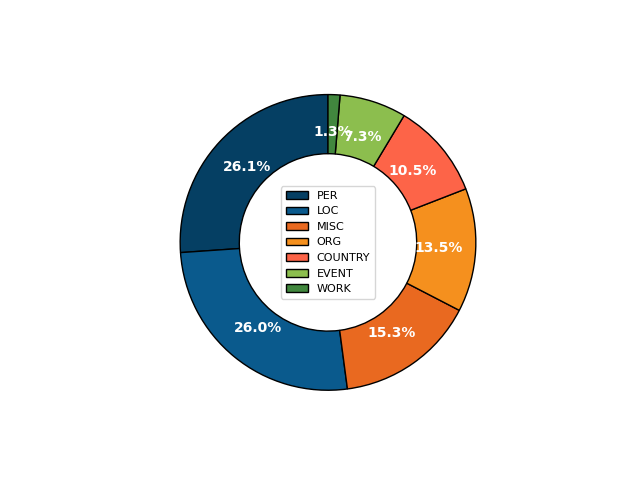}
  \caption{Entity type distribution of WIKIDiverse.}
  \label{wikid_entity_type}
\end{figure}

\subsection{WikiDiverse and WikiMEL}
\label{WikiDiverse}
WikiDiverse is a high-quality human-annotated MEL dataset with diversified contextual topics and entity types from Wikinews, which uses Wikipedia as the corresponding knowledge base. WikiMEL is collected from Wikipedia entities pages and contains more than 22k multimodal sentences. The statistics of WIKIDiverse and WikiMEL are shown in Table \ref{tab:WIKIDiverse}. The entity type distribution of WIKIDiverse is illustrated in Figure \ref{wikid_entity_type}. 

During the data collection process, we select the entire WIKIDiverse dataset along with 5,000 samples from the WikiMEL dataset. Compared to WikiMEL, WIKIDiverse features more content-rich images that better represent real-world application scenarios, making it particularly suitable for meeting the requirements of the VP-MEL task in practical contexts. Consequently, WIKIDiverse constitutes the majority of the VPWiki dataset. Additionally, we integrate the knowledge bases (KBs) from both datasets, resulting in an entity set encompassing all entities in the main namespace.

\begin{table}[!]
\small
\setlength\tabcolsep{4pt}
\renewcommand{\arraystretch}{1.3}
\centering
\begin{tabular}{lccccccc}
\toprule
\multirow{2}{*}{\textbf{Methods}}
 & \multicolumn{3}{|c|}{\textbf{WikiMEL}}  & \multicolumn{3}{c}{\textbf{RichpediaMEL}}  \\
\cline{2-7}
 & \multicolumn{1}{|c}{\textbf{H@1}} & \textbf{H@3} & \multicolumn{1}{c|}{\textbf{H@5}} & \textbf{H@1} & \textbf{H@3} & \multicolumn{1}{c}{\textbf{H@5}} \\
 \midrule 
 ViLT$^*$ & \multicolumn{1}{|c}{ 72.64} &  84.51  & \multicolumn{1}{c|}{ 87.86} &45.85 &62.96 & 69.80\\
 ALBEF$^*$ & \multicolumn{1}{|c}{78.64} & 88.93 & \multicolumn{1}{c|}{91.75} &65.17 & 82.84 & 88.28\\
 CLIP$^*$ & \multicolumn{1}{|c}{83.23 } &92.10  & \multicolumn{1}{c|}{94.51 } &67.78& 85.22 &90.04\\
 METER$^*$ & \multicolumn{1}{|c}{72.46} & 84.41& \multicolumn{1}{c|}{ 88.17} &63.96&82.24 & 87.08\\
 BERT$^*$ & \multicolumn{1}{|c}{74.82} &86.79 & \multicolumn{1}{c|}{90.47} &59.55 &81.12 & 87.16\\
 BLINK$^*$ & \multicolumn{1}{|c}{74.66} & 86.63& \multicolumn{1}{c|}{ 90.57} &58.47&81.51& 88.09\\
 JMEL$^*$ & \multicolumn{1}{|c}{64.65} & 79.99 & \multicolumn{1}{c|}{84.34} &48.82 & 66.77 & 73.99\\
 VELML & \multicolumn{1}{|c}{68.90} &83.50 & \multicolumn{1}{c|}{87.77} &62.80 &82.04 &87.84\\
 GHMFC & \multicolumn{1}{|c}{75.54} &88.82 & \multicolumn{1}{c|}{92.59} &76.95 &88.85& 92.11\\

 MIMIC & \multicolumn{1}{|c}{87.98} &95.07 & \multicolumn{1}{c|}{96.37} & 81.02 & 91.77 & 94.38\\
 MELOV$^*$ & \multicolumn{1}{|c}{88.91} & 95.61 & \multicolumn{1}{c|}{96.58} & 84.14 &92.81 & 94.89\\
  \midrule 

      \rowcolor[gray]{0.92} IIER & \multicolumn{1}{|c}{\textbf{88.93}} & \textbf{95.69}& \multicolumn{1}{c|}{\textbf{96.73}} & \textbf{84.63}  & \textbf{93.27} &\multicolumn{1}{c}{\textbf{95.30}}\\
     \midrule 
 
\end{tabular}
\caption{Baseline results marked with "$*$" according to \citet{sui-etal-2024-melov}. We run each method three times with different random seeds and
 report the mean value of every metric. The best score is highlighted in bold.}
\label{MoreMEL}
\end{table}

\subsection{Data Samples}
\label{moresamples}
We provide additional data samples categorized by entity type. The specific details can be found in Figure \ref{fig:datasamples}.

\subsection{Additional Experiments}
\label{Experiments}
To comprehensively assess the performance of the IIER framework in the MEL task, we test IIER on the WikiMEL and RichpediaMEL datasets \citep{10.1145/3477495.3531867}. The statistics of WikiMEL and RichpediaMEL are shown in Table \ref{tab:WIKIDiverse}. Experimental results are shown in Table \ref{MoreMEL}. 
The experimental results demonstrate that IIER remains highly competitive with state-of-the-art MEL method. 

It is noted that within these two datasets, certain metrics of IIER exhibit values that are comparable to those of MELOV, such as H@1 and H@3 in the WikiMEL dataset. This may be attributed to the higher image quality and the homogeneous entity types (primarily $Person$) in WikiMEL and RichpediaMEL. 
When datasets contain fewer entity types and minimal image noise, the auxiliary information generated by IIER contributes less to performance improvement.

Nevertheless, IIER achieves the best performance on WikiDiverse, which includes a wider variety of entity types, and achieves a new SOTA for the VP-MEL task. As MEL increasingly addresses more complex scenarios, IIER shows significant potential for future advancements.

\begin{figure*}[h!]
    \centering
    \begin{minipage}{\textwidth}
        \centering
        \begin{subfigure}[b]{0.9\textwidth} 
            \centering
            \includegraphics[width=0.9\textwidth]{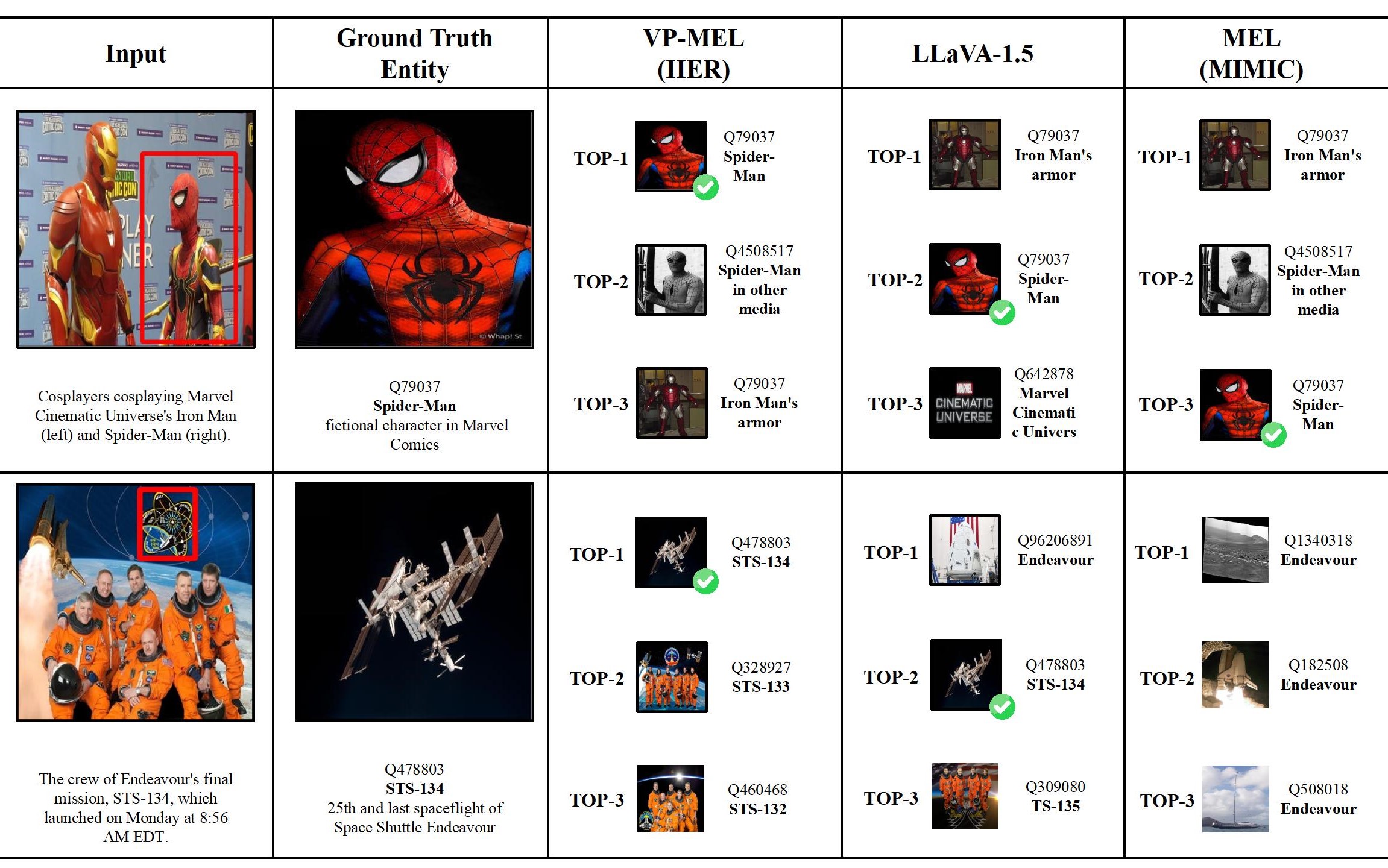}
            \caption{Successful predictions.}
            \label{Successful predictions}
        \end{subfigure}
        \vspace{1em} 
        \begin{subfigure}[b]{0.9\textwidth} 
            \centering
            \includegraphics[width=0.9\textwidth]{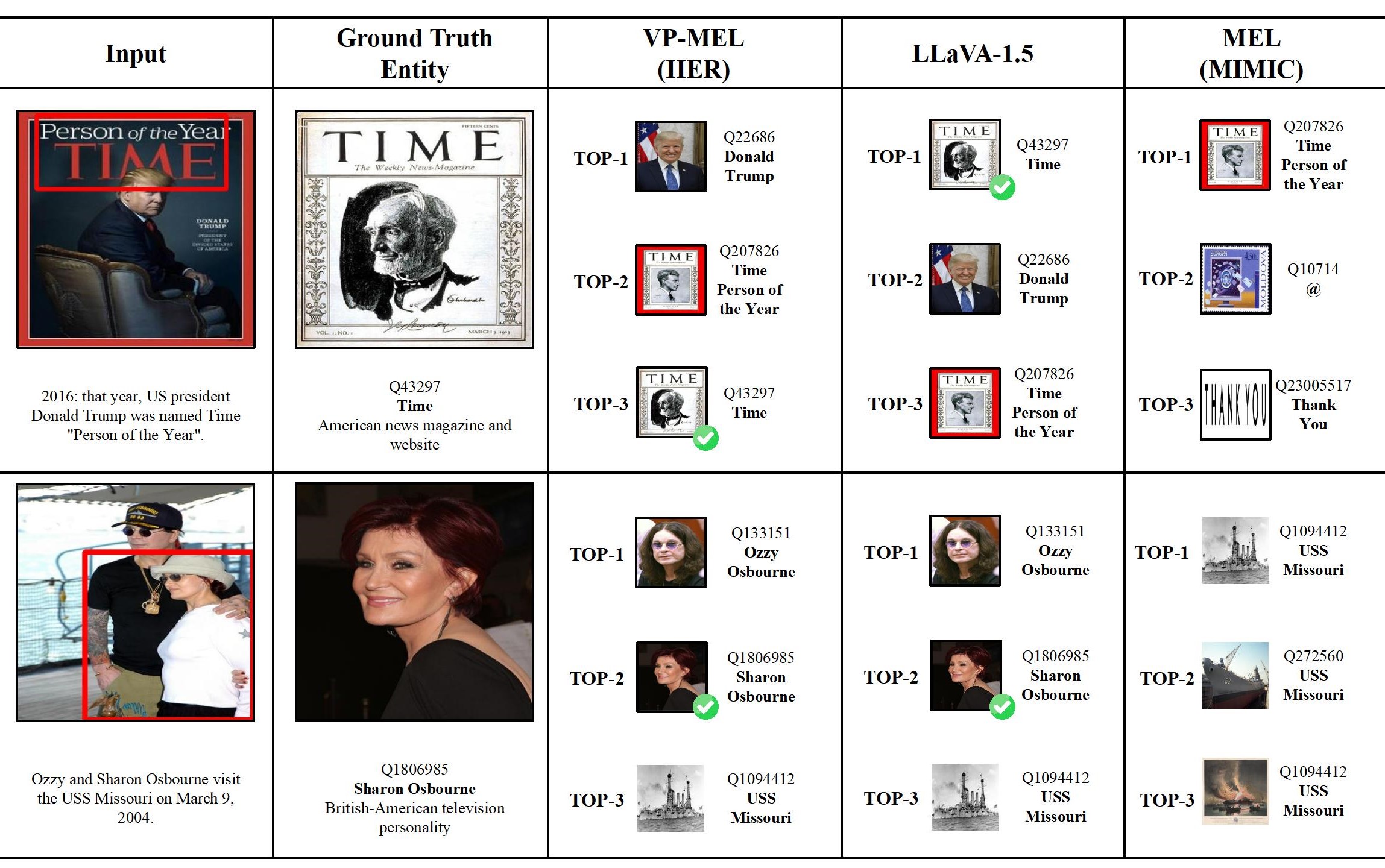}
            \caption{Failed predictions.}
            \label{Failed predictions}
        \end{subfigure}
    \end{minipage}
    \caption{Case study for VP-MEL. Each row is a case, which contains Input, ground truth entity, and top three retrieved entities of three methods, \emph{i.e.}, IIER (ours), LLaVA-1.5 \citep{Liu_2024_CVPR}, MIMIC \citep{10.1145/3580305.3599439}. Each retrieved entity is described by its Wikidata QID and entity name, with the entity marked with a checkmark indicating the correct one.}
    \label{case_study}
\end{figure*}

\begin{figure*}[t!]
	\centering
	\includegraphics[scale=0.060 ]{datasamples.jpg}
	\caption{Examples of the VPWiki dataset. Each row represents a sample corresponding to a specific entity type, which contains the entity type, image of mention, text of mention, image of entity, and entity.}
	\label{fig:datasamples}
\end{figure*}

\subsection{Case Study}
\label{Case}
To clearly demonstrate the proposed VP-MEL task and the IIER model, we conduct case studies and compare them against two strong competitors (\emph{i.e.}, LLaVA-1.5 and MIMIC), in Figure \ref{case_study}. As shown in Figure \ref{Successful predictions}, in the first case, all three methods correctly predicted the entity. IIER makes full use of both image and text information, allowing it to more effectively distinguish between the different individuals in the image. LLaVA-1.5 may be overwhelmed by the textual information, while MIMIC struggles to identify the correct entity when the mention words are unavailable. In the second case, both LLaVA-1.5 and MIMIC retrieve \textit{Endeavour} as the first choice. Only IIER, with the guidance of Visual Prompts and integration of textual information, correctly predicts the right entity. In Figure \ref{Failed predictions}, we present the failed predictions. In the first case, when the content of the image interferes with the visual prompt, it impairs the reasoning process of IIER. The red box in the image bears a high similarity to the visual prompt. As a result, IIER incorrectly focuses on the wrong region of the image, ranking \textit{Donald Trump} first. When IIER encounters difficulties in distinguishing the objects within the visual prompts, it leads to incorrect inferences. For example, in the second case, the distinguishing features of the two individuals in the image are obstructed, which causes IIER to struggle in differentiating between them. The image content in real-world data is often complex, which makes VP-MEL a challenging task. We hope that this task can be further refined and developed over time.

\end{document}

%% file: math_commands.tex
\usepackage{amsmath,amsfonts,bm}

\def\eqref#1{equation~\ref{#1}}

\def\1{\bm{1}}

\DeclareMathAlphabet{\mathsfit}{\encodingdefault}{\sfdefault}{m}{sl}
\SetMathAlphabet{\mathsfit}{bold}{\encodingdefault}{\sfdefault}{bx}{n}

